\documentclass[twoside,11pt]{article}
\usepackage{jair, theapa, rawfonts}
\usepackage{graphicx}
\usepackage{mathtools}
\usepackage{amsfonts}       
\usepackage{amsmath}
\usepackage{nicefrac}       
\usepackage{microtype}      
\usepackage{xspace}

\usepackage{times}
\usepackage{epsfig}
\usepackage{amssymb}
\usepackage{bbm}
\usepackage{multirow}
\usepackage{xcolor}
\usepackage{url}

\jairheading{1}{2019}{}{}{}
\ShortHeadings{}
{}
\firstpageno{1}

\DeclareMathOperator{\softmax}{softmax}
\DeclareMathOperator{\entropy}{Entropy}

\DeclareMathOperator{\LSTM}{LSTM}
\DeclareMathOperator{\MLP}{MLP}

\newcommand{\etal}{{et al.}\xspace}
\newcommand{\FaceAttend}{\textsc{Face-Attend}\xspace}
\newcommand{\FaceAttendF}{\textsc{Dual-Face-Att}\xspace}
\newcommand{\FaceAttendL}{\textsc{Joint-Face-Att}\xspace}
\newcommand{\FaceCap}{\textsc{Face-Cap}\xspace}
\newcommand{\FaceCapL}{\textsc{Face-Cap-Memory}\xspace}
\newcommand{\FaceCapF}{\textsc{Face-Cap-Repeat}\xspace}
\newcommand{\FaceInit}{\textsc{Init-Flow}\xspace}
\newcommand{\FaceStep}{\textsc{Step-Inject}\xspace}
\newcommand{\XU}{\textsc{Show-Att-Tell}\xspace}
\newcommand{\ANDERSON}{\textsc{Up-Down}\xspace}
\newcommand{\FER}{\textsc{FER-2013}\xspace}

\newcommand{\green}[1]{\textcolor{black}{#1}}

\begin{document}

\title{Image Captioning using Facial Expression and Attention}

\author{  \name Omid Mohamad Nezami \email omid.mohamad-nezami@hdr.mq.edu.au \\
          \addr Macquarie University, Sydney, NSW, Australia \\
                CSIRO's Data61, Sydney, NSW, Australia
          \AND
          \name Mark Dras \email mark.dras@mq.edu.au \\
          \addr Macquarie University, Sydney, NSW, Australia
          \AND
          \name Stephen Wan \email stephen.wan@data61.csiro.au \\
          \addr CSIRO's Data61, Sydney, NSW, Australia
          \AND
          \name C\'ecile Paris \email cecile.paris@data61.csiro.au \\
          \addr Macquarie University, Sydney, NSW, Australia \\
                CSIRO's Data61, Sydney, NSW, Australia
      }


\maketitle

\begin{abstract}
Benefiting from advances in machine vision and natural language processing techniques, current image captioning systems are able to generate detailed visual descriptions. 
For the most part, these descriptions represent an objective characterisation of the image, although some models do incorporate subjective aspects related to the observer's view of the image, such as sentiment;
current models, however, usually do not consider the emotional content of images during the caption generation process. This paper addresses this issue by proposing novel image captioning models which use facial expression features to generate image captions. The models generate image captions using long short-term memory networks applying facial features in addition to other visual features at different time steps. We compare a comprehensive collection of image captioning models with and without facial features using all standard evaluation metrics. The evaluation metrics indicate that applying facial features with an attention mechanism achieves the best performance, showing more expressive and more correlated image captions, on an image caption dataset extracted from the standard Flickr 30K dataset, consisting of around 11K images containing faces. An analysis of the generated captions finds that, perhaps unexpectedly, the improvement in caption quality appears to come not from the addition of adjectives linked to emotional aspects of the images, but from more variety in the actions described in the captions.
\end{abstract}

\section{Introduction}
\label{Introduction}

Image captioning systems aim to describe the content of an image using Computer Vision and Natural Language Processing approaches which have led to important and practical applications such as helping visually impaired individuals \cite{vinyals2015show}. This is a challenging task because we have to capture not only the objects but also their relationships, and the activities displayed in the image in order to generate a meaningful description. The impressive progress of deep neural networks and large image captioning datasets has resulted in a considerable improvement in generating automatic image captions \cite{vinyals2015show,xu2015show,johnson2016densecap,you2016image,rennie2017self,chen2017sca,lu2017knowing,anderson2018bottom,tian2019multi}.

\begin{figure}
  \centering
  \includegraphics[width=1.0\textwidth]{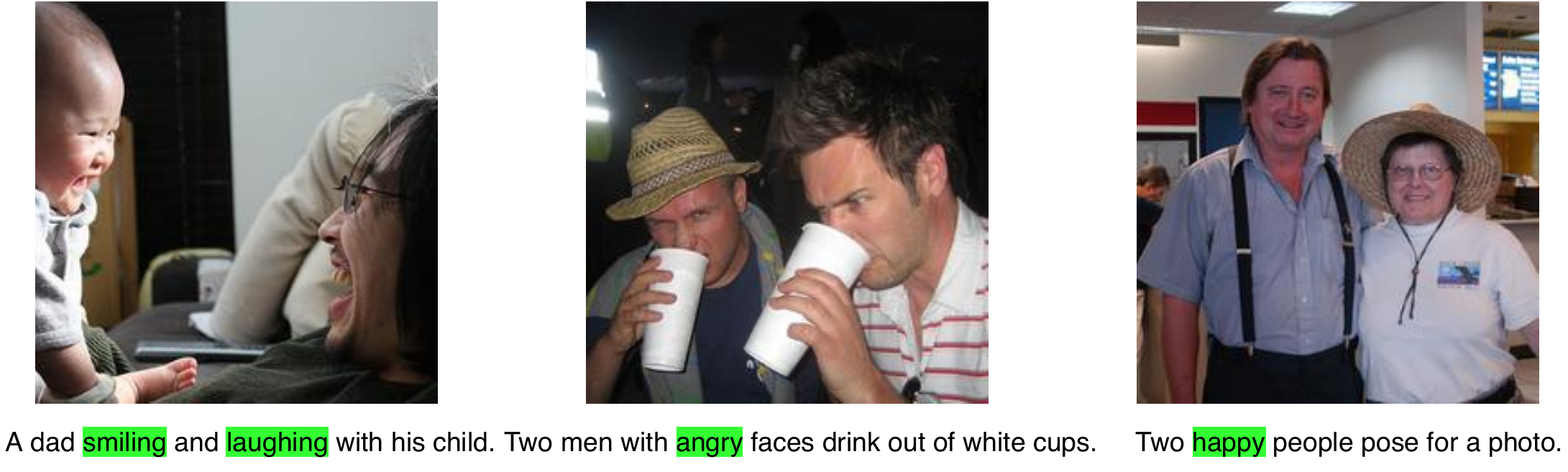}
  \caption{The examples of Flickr 30K dataset~\cite{young2014image} with emotional content. The green color indicates words with strong emotional values.}
  \label{fig:example}
\end{figure}

However, current image captioning methods often overlook the emotional aspects of the image, which play an important role in generating captions that are more semantically correlated with the visual content. For example, Figure~\ref{fig:example} shows three images with their corresponding human-generated captions including emotional content. The first image at left has the caption of ``a dad smiling and laughing with his child'' using ``smiling'' and ``laughing'' to describe the emotional content of the image. In a similar fashion, `angry'' and ``happy'' are applied in the second and the third images, respectively. These examples demonstrate how image captioning systems that recognize emotions and apply them can generate richer, more expressive and more human-like captions; this idea of incorporating emotional content is in fact one that is typical to intelligent systems, which researchers like \citeA{lisetti1998affective} have identified as necessary to generate more effective and adaptive outcomes. 
Although detecting emotions from visual data has been an active area of research in recent years~\cite{fasel2003automatic,sariyanidi2015automatic}, designing an effective image captioning system to employ emotions in describing an image is still an open and challenging problem.

A few models have incorporated sentiment or other non-factual information into image captions~\cite{gan2017stylenet,mathews2016senticap,chen2018factual}; 
they typically require the collection of a supplementary dataset, from which a sentiment vocabulary is derived, drawing on work in Natural Language Processing \cite{pang-lee:2008} where sentiment is usually characterized as one of positive, neutral or negative. \citeA{mathews2016senticap}, for instance, constructed a sentiment image-caption dataset via crowdsourcing, where annotators were asked to include either positive sentiment (e.g. \textit{a cuddly cat}) or negative sentiment (e.g. \textit{a sinister cat}) using a fixed vocabulary; their model was trained on both this and a standard set of factual captions. These kinds of approaches typically embody descriptions of an image that represent an \textit{observer's} view towards the image (e.g. \textit{a cuddly cat} for a positive view of an image, versus \textit{a sinister cat} for a negative one); they do not aim to capture the emotional content of the image, as in   Figure~\ref{fig:example}.

To capture the emotional content of the image, we propose two groups of models: \FaceCap \footnote{An earlier version of \FaceCap has already been published \cite{nezami2018face}.} and \FaceAttend. \FaceCap feeds in a fixed one-hot encoding vector similar to \citeA{hu2017toward} and \citeA{you2018image}. In comparison, we represent the aggregate facial expressions of the input image at different time steps of our caption generator, which employs a long short-term memory (LSTM) architecture. To construct the vector, we train a state-of-the-art facial expression recognition (FER) model which automatically recognizes facial expressions (e.g. happiness, sadness, fear, and so on). However, the recognized facial expressions are not always reliable because the FER model is not 100\% accurate. This can result in an image captioning architecture that propagates errors.
Moreover, these facial expression classes do not necessarily align with more fine-grained facial expression representations such as action units (AUs), one framework for characterising different facial muscle movements \cite{lucey2010extended}. Hence, we propose an alternative representation that uses more fine-grained facial expression features (e.g. convolutional features) which could potentially be more useful than the one-hot encoding representation.
We also recognize from design choices that there might be images that \FaceCap may not perform well on (e.g. images including multiple faces such as Figure~\ref{fig:example}, because we have a single encoding representation of emotion for the whole image) and an attention mechanism might better localise emotional features in a way useful for image captioning.
Thus, \FaceAttend employs an attention mechanism to selectively attend to facial features, for different detected faces in an image, extracted from the last convolutional layer of the FER model. 
\FaceAttend uses two LSTMs to incorporate facial features along with general visual content in generating image descriptions. 

The main contributions of the paper are highlighted as follows:






\begin{itemize}
    \item We propose \FaceCap and \FaceAttend models to effectively employ facial expression features with general visual content to generate image captions. To the authors' knowledge, this is the first study to apply facial expression analyses in image captioning tasks.
    
	\item Our generated captions using the models are evaluated by all standard image captioning metrics. The results show the effectiveness of the models comparing to a comprehensive list of image captioning models using the FlickrFace11K dataset,\footnote{Our dataset splits and labels are publicly available: \url{https://github.com/omidmnezami/Face-Cap}} the subset of images from the Flickr 30K dataset~\cite{young2014image} that include human faces.
    
    \item We further assess the quality of the generated captions in terms of the characteristics of the language used, such as variety of expression. Our analysis suggests that the generated captions by our models improve over other image captioning models by better describing the actions performed in the image.
    
\end{itemize}

\section{Previous Work}
In the following sections, we review image captioning and facial expression recognition models as they are the key parts of our work.

\subsection{Image Captioning}

There are three main types of image captioning systems: template-based models, retrieval-based models and deep-learning based models \cite{bernardi2016automatic,hossain2019comprehensive}. Template-based ones first detect visual objects, their attributes and relations and then fill a pre-defined template's blank slots \cite{farhadi2010every}. Retrieval-based ones generate captions using the available captions corresponding to similar images in their corresponding datasets \cite{hodosh2013framing}. These classical image captioning models have some limitations. For example, template-based ones cannot generate a wide variety of captions with different lengths, and retrieval-based ones are not able to generate specifically-designed captions for different images. Moreover, classical models do not incorporate the detection and generation steps using an end-to-end training approach. Because of these limitations, modern image captioning models using deep learning are currently the most popular.

Modern image captioning models usually use an encoder-decoder paradigm~\cite{kiros2014unifying,vinyals2015show,xu2015show}. They apply a top-down approach where a Convolutional Neural Network (CNN) model learns the image content (encoding), followed by a Long Short-Term Memory (LSTM) generating the image caption (decoding). This follows the paradigm employed in machine translation tasks, using deep neural networks~\cite{sutskever2014sequence}, to translate an image into a caption. 
This top-down mechanism directly converts the extracted visual features into image captions \cite{chen2015mind,donahue2015long,johnson2016densecap,karpathy2015deep,mao2014deep}. However, attending to fine-grained and important fragments of visual data in order to provide a better image description is usually difficult using a top-down paradigm. To solve this problem, a combination of a top-down approach and a bottom-up approach, inspired from the classical image captioning models, is proposed by \citeA{you2016image}. The bottom-up approach overcomes this limitation by generating the relevant words and phrases, which can be detected from visual data with any image resolution, and combining them to form an image caption~\cite{elliott2013image,farhadi2010every,kulkarni2013babytalk,kuznetsova2012collective}.

To attend to fine-grained fragments, attention-based image captioning models have been recently proposed \cite{xu2015show}. These kinds of approaches usually analyze different regions of an image in different time steps of a caption generation process, in comparison to the initial encoder-decoder image captioning systems which consider only the whole image \cite{vinyals2015neural} as an initial state for generating image captions. They can also take the spatial information of an image into account to generate the relevant words and phrases in the image caption. The current state-of-the-art models in image captioning are attention-based systems~\cite{anderson2018bottom,rennie2017self,xu2015show,you2016image}, explained in the next section, similar to our attention-based image captioning systems.

\subsubsection{Image Captioning with Attention}
\label{sec:rel_att}

Visual attention is an important aspect of the visual processing system of humans \cite{koch1987shifts,corbetta2002control,spratling2004feedback,rensink2000dynamic}. It dynamically attends to salient spatial locations in an image with special properties or attributes which are relevant to particular objects. It is different from dealing with the whole image as a set of static extracted features, and assists humans to concentrate more on a targeted object or region at each time step. Although visual attention has been extensively studied in Psychology and Neuroscience \cite{desimone1995neural,eriksen1986visual,coffman2014battery}, it has only more recently been adapted to different artificial intelligence fields including machine learning, computer vision and natural language processing.

The first image captioning model with attention was proposed by \citeA{xu2015show}. The model uses visual content extracted from the convolutional layers of CNNs, referred to as spatial features, as the input of a \emph{spatial} attention mechanism to selectively attend to different parts of an image at every time step in generating an image caption. This work is inspired by the work of \citeA{bahdanau2014neural}, since extended by \citeA{vaswani2017attention}, who employed attention in the task of machine translation; by \citeA{mnih2014recurrent}; and by \citeA{ba2014multiple} who applied attention in the task of object recognition. Image captioning with attention differs from previous encoder-decoder image captioning models by concentrating on the salient parts of an input image to generate its equivalent words or phrases simultaneously. \citeA{xu2015show} proposed two types of attention including a hard (stochastic) mechanism and a soft (deterministic) mechanism. In the soft attention mechanism, a weighted matrix is calculated to weight a particular part of an image as the input to the decoder (interpreted as a probability value for considering the particular part of the image). The hard attention mechanism, in contrast, picks a sampled annotation vector corresponding to a particular part of an image at each time step as the input to the decoder.

\citeA{rennie2017self} extended the work of Xu~\etal by using the CIDEr metric~\cite{vedantam2015cider}, a standard performance metric for image captioning, to optimize their caption generator instead of optimizing maximum likelihood estimation loss. Their approach was inspired by a Reinforcement Learning approach \cite{williams1992simple,sutton1998introduction} called self-critical sequence training, which involves normalizing the reward signals calculated using the CIDEr metric at test time.

\citeA{yu2017end} and \citeA{you2016image} applied a notion of \emph{semantic} attention to detected visual attributes, learned in an end-to-end fashion, where bottom-up approaches were combined with top-down ones to take advantage of both paradigms. For instance, they acquired a list of semantic concepts or attributes, regarded as a bottom-up mechanism, and used the list with visual features, as an instance of top-down information, to generate an image caption. Semantic attention is used to attend to semantic concepts detected from various parts of a given image. Here, the visual content was only used in the initial time step. In other time steps, semantic attention was used to select the extracted semantic concepts. That is, semantic attention differs from spatial attention, which attends to spatial features in every time step, and does not preserve the spatial information of the detected concepts.

To preserve spatial information, salient regions can be localized using spatial transformer networks~\cite{jaderberg2015spatial}, which get the spatial features as inputs. This is similar to Faster R-CNN's generation of bounding boxes~\cite{ren2017faster}, but it is trained in an end-to-end fashion using bilinear interpolation instead of a Region of Interest pooling mechanism as proposed by~\citeA{johnson2016densecap}. Drawing on this idea, \citeA{anderson2018bottom} applied spatial features to image captioning by using a pre-trained Faster R-CNN and an attention mechanism to discriminate among different visual-based regions regarding the spatial features. Specifically, they combined bottom-up and top-down approaches where a pre-trained Faster R-CNN is used to extract the salient regions from images, instead of using the detected objects 
as high-level semantic concepts in the work of \citeA{you2016image}; and an attention mechanism is used to generate spatial attention weights over the convolutional feature maps representing the regions. Faster R-CNN, as an object detection model, is pre-trained on the Visual Genome dataset \cite{krishna2017visual}; this pre-training on a large dataset is analogous to pre-training a classification model on the ImageNet dataset \cite{russakovsky2015imagenet}. \citeA{jin2015aligning} previously used salient regions with different scales which are extracted by applying selective search \cite{uijlings2013selective} instead of applying Faster R-CNN. Then, they made the input of their spatial attention mechanism by resizing and encoding the regions in the task of image captioning.


In our image captioning systems, we use an attention mechanism weighting visual features as a top-down approach. We also use another attention mechanism to attend to facial expression features as a bottom-up approach. This combination allows our image captioning models to generate captions which are highly correlated with visual content and facial features. To do so, we train a state-of-the-art facial expression recognition model to extract the features. Then, we use the features, attended using the attention mechanism at each time step, to enrich image captions by targeting emotional values.

\subsubsection{Image Captioning with Style}
\label{sec:rel_st}

Most image captioning systems concentrate on describing objective visual content without adding any extra information, giving rise to factual linguistic descriptions. However, there are also stylistic aspects of language which play an essential role in enriching written communication and engaging users during interactions. Style helps in clearly conveying visual content \cite{mathews2018semstyle}, and making the content more attractive \cite{gan2017stylenet,chen2018factual}. It also conveys personality-based \cite{pennebaker1999linguistic} and emotion-based attributes which can impact on decision making \cite{mathews2016senticap}. Incorporating style into the description of an image is effective in boosting the engagement level of humans with respect to dialogue in visually-grounded chatbot platforms \cite{huber2018emotional} and in interacting with automatically-generated comments for photos and videos in social media platforms \cite{li2016share}.

There are a few models that have incorporated style or other non-factual characteristics into the generated captions~\cite{mathews2016senticap,gan2017stylenet,nezami2018senti,nezami2019towards}. In addition to describing the visual content, these models learn to generate different forms or styles of captions. For instance, \citeA{mathews2016senticap} proposed the Senti-Cap system to generate sentiment-bearing captions. Here, the notion of sentiment is drawn from Natural Language Processing~\cite{pang-lee:2008}, with sentiment either \textit{negative} or \textit{positive}. The Senti-Cap system of \citeA{mathews2016senticap} is a full switching architecture incorporating both factual and sentiment caption paths. In comparison, the work of \citeA{gan2017stylenet} consists of a Factored-LSTM learning the stylistic information in addition to the factual information of the input captions. \citeA{chen2018factual} subsequently applied a mechanism to weight the stylistic and the factual information using Factored-LSTM. All these approaches need two-stage training: training on factual image captions and training on sentiment-bearing image captions. Therefore, they do not support end-to-end training.

To address this issue, \citeA{you2018image} designed two new schemes, Direct Inject and Sentiment Flow, to better employ sentiment in generating image captions. For Direct Inject, an additional dimension was added to the input of a recurrent neural network (RNN) to express sentiment,\footnote{A related idea was earlier proposed by \citeA{radford2017learning} who identified a sentiment unit in a RNN-based system.} and the sentiment unit is injected at every time step of the generation process. The Sentiment Flow approach of \citeA{you2018image} injects the sentiment unit only at the initial time step of a designated sentiment cell trained in a similar learning fashion to the memory cell in LSTMs. 

All of the above work is focused on subjective descriptions of images using a given sentiment vocabulary, rather than representing the emotional content of the image, as we do in this work.
In order to target content-based emotions using visual data, we propose \FaceCap and \FaceAttend models employing attention mechanisms to attend to visual features. We aim to apply the emotional content, recognized using a facial expression analysis, of images themselves during a caption generation process. We use the emotional content to generate image captions without any extra style-based or sentiment-bearing vocabulary: our goal is to see whether, given the existing vocabulary, incorporating the emotional content can produce better captions.

\subsection{Facial Expression Recognition}

\begin{figure}
    \begin{center}
    \includegraphics[width=0.62\linewidth]{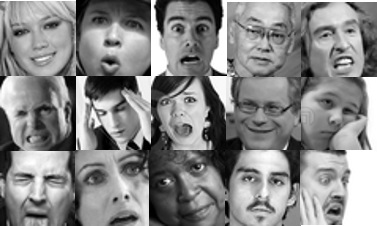}
    \end{center}
    \caption{Examples from the Facial Expression Recognition 2013 dataset~\cite{goodfellow2013challenges} including seven standard facial expressions.}
\label{fig:fer_samples}
\end{figure}

Facial expression is a form of non-verbal communication conveying attitudes, affects, and intentions of individuals. It happens as the result of changes over time in facial features and muscles~\cite{fasel2003automatic}. It is also one of the most important communication means for showing emotions and transferring attitudes in human interactions. Indeed, research on facial expressions started more than a century ago when Darwin published his book titled, ``The expression of the emotions in man and animals''~\cite{ekman2006darwin}. Since then a large body of work has emerged on recognizing facial expressions, usually using a purportedly universal framework of a small number of standard emotions (\textit{happiness}, \textit{sadness}, \textit{fear}, \textit{surprise}, \textit{anger}, and \textit{disgust}) or this set including a \textit{neutral} expression~\cite{field1982discrimination,kanade2000comprehensive,fasel2003automatic,yin20063d,fridlund2014human,sariyanidi2015automatic,nezami2019shemo} or more fine-grained facial features such as facial action units, defined as the deformations of facial muscles \cite{tian2001recognizing}. Recently, recognizing facial expressions has been paid special attention because of its practical applications in different domains such as education \cite{nezami2017semi,nezami2018deep}, health-care and virtual reality \cite{zeng2008survey,fasel2003automatic}. It is worth mentioning that the automatic recognition of facial expressions is a difficult task because different people express their attitudes in different ways and there are close similarities among various types of facial expressions \cite{zeng2018facial} as shown in Figure \ref{fig:fer_samples}.

To find effective representations, deep learning based methods have been recently successful in this domain. Due to their complex architectures including multiple layers, they can capture hierarchical structures from low- to high-level representations of facial expression data. \citeA{tang2013deep}, the winner of the 2013 Facial Expression Recognition (FER) challenge~\cite{goodfellow2013challenges}, trained a Convolutional Neural Network (CNN) with a linear support vector machine (SVM) to detect facial expressions. He replaced the softmax layer, used to generate a probability distribution across multiple classes, with a linear SVM and showed a consistent improvement compared to the previous work. Instead of cross-entropy loss, his approach optimizes a margin-based loss to maximize margins among data points belonging to diverse classes.

CNNs are also used for feature extraction and transfer learning in this domain. \citeA{kahou2016emonets} applied a CNN model to recognize facial expressions. Their approach uses a combination of deep neural networks to learn from diverse data modalities including video frames, audio data and spatio-temporal information. The CNN model, as the best model in this work, aims to recognize emotions from static video frames. Then the recognized emotions are combined across a video clip by a frame aggregation technique and classified using an SVM with a radial basis function kernel. \citeA{yu2015image} used an ensemble of CNNs to detect facial expressions in a transfer learning framework. On their target samples, they applied a set of face detection approaches to optimally detect faces and remove irrelevant data. They used a multiple neural network training framework to learn a set of weights assigned to the responses of the CNNs in addition to averaging and voting over the responses. \citeA{kim2016fusing} combined aligned and non-aligned faces to enhance the recognition performance of facial expressions where they automatically detected facial landmarks from faces to rotate and align faces. Then, they trained a CNN model using this combination of faces. \citeA{zhang2015learning} proposed a CNN-based method to recognize social relation traits (e.g. friendly, competitive and dominant) from detected faces in an image. The method includes a CNN model to recognize facial expressions projected into a shared representation space. The space combines the extracted features from two detected faces in an image and generates the predictions of social traits.

The models mentioned above usually use conventional CNN architectures to report the performance on different facial expression recognition datasets including the FER-2013 dataset~\cite{goodfellow2013challenges}, which is a publicly available dataset with a large number of human faces captured in real-life settings. \citeA{pramerdorfer2016facial} instead used an ensemble of very deep architectures of CNNs such as VGGnet, Inception and ResNet by identifying the bottlenecks of the previous state-of-the-art facial expression recognition models on the FER-2013 dataset and achieving a new state-of-the-art result on the dataset. The quality of these recent models is high: it is at least as good as human performance~\cite{goodfellow2013challenges}.
The idea of applying VGGnet in facial expression recognition tasks motivates our work to make a facial expression recognition module reproducing the state-of-the-art result on FER-2013 dataset. We use the module to extract facial features from human faces to apply in our image captioning models.


\section{Approach}
In this section, we describe \FaceCap and \FaceAttend, our proposed models for generating image captions using facial expression analyses. The models are inspired by two popular image captioning models, specifically Show-Attend-Tell \cite{xu2015show} and Up-Down-Captioner \cite{anderson2018bottom}.

Show-Attend-Tell is a well-known and widely used image captioning system that incorporates an attention mechanism to attend to spatial visual features. It demonstrates a significant improvement over earlier image captaining models that do not have an attention mechanism. From this starting point, we propose the \FaceCap model which similarly attends to visual features and additionally uses facial expression analyses in generating image captions. \FaceCap incorporates a one-hot encoding vector as a representation of the facial expression analysis, similar to the representations used for sentiment by \citeA{hu2017toward} and \citeA{you2018image}. 

Up-Down-Captioner is a current state-of-the-art image captioning model, defining a new architecture to incorporate attended visual features in generating image captions. In this model, the features directly relate to the objects in the image and two LSTMs (one for generating attention weights and another one for a language model) are used to generate image captions. We propose \FaceAttend based on this kind of architecture, as we can apply more fine-grained facial expression features and use two LSTMs to attend to the features in addition to the general visual features.
Because Up-Down-Captioner already incorporates attention on objects in the image, our models derived from this allow us to examine the effectiveness of the facial expression features beyond just recognition of the face as an object.

In what follows, we describe our datasets and our facial expression recognition model that are used by \FaceCap and \FaceAttend. We then explain \FaceCap in Section \ref{sec:faceCap} and \FaceAttend in Section \ref{sec:faceAttend}.

\subsection{Datasets}
\label{sec:datasets}

\paragraph{Facial Expression Recognition}
To train our facial expression recognition model, we use the facial expression recognition 2013 (\FER) dataset~\cite{goodfellow2013challenges}. It includes images labeled with standard facial expression categories (\textit{happiness}, \textit{sadness}, \textit{fear}, \textit{surprise}, \textit{anger}, \textit{disgust} and \textit{neutral}). It consists of 35,887 examples (standard splits are 28,709 for training, 3589 for public and 3589 for private test), collected by means of the Google search API. The examples are in grayscale at the size of 48-by-48 pixels. For our purposes, we split the standard training set of \FER into two sections after removing 11 completely black examples: 25,109 for training our models and 3589 for development and validation. Similar to other work in this domain~\cite{kim2016fusing,pramerdorfer2016facial,yu2015image}, we use the private test set of FER-2013 for the performance evaluation of the model after the training phase. To compare with the related work, we do not apply the public test set either for training or for validating the model.

\paragraph{Image Captioning}
To train \FaceCap and \FaceAttend, we have extracted a subset of the Flickr 30K dataset with image captions~\cite{young2014image} that we name FlickrFace11K. It contains 11,696 images including human faces detected using a convolutional neural network-based face detector~\cite{king2009dlib}.\footnote{The new version (2018) of Dlib library is applied.} Each image has five ground-truth captions. We observe that the Flickr 30K dataset is a good source for our dataset, because it has a larger portion of images that include human faces, in comparison with other image caption datasets such as the MSCOCO dataset~\cite{lin2014microsoft}. We split the FlickrFace11K samples into 8696 for training, 2000 for validation and 1000 for testing. Since we aim to train a facial expression recognition model on FER-2013 and use it as a facial expression feature extractor on the samples of FlickrFace11K, we need to make the samples consistent with the FER-2013 data. To this end, the face detector is used to pre-process the faces of FlickrFace11K. The faces are cropped from each sample. Then, we transform each face to grayscale and resize it into 48-by-48 pixels, which is the same as in the FER-2013 data.


\subsection{Facial Expression Recognition Model}
\label{sec:FER_model}
For our core models, we train a facial expression recognition (FER) model using the VGG-B architecture~\cite{simonyan2014very}, because of its strong performance in \citeA{pramerdorfer2016facial}.  We remove the last convolutional block, including two convolutional layers, and the last max pooling layer from the architecture. We use $3\times3$ kernel sizes for all remaining convolutional layers. We use a batch normalization layer \cite{ioffe2015batch} after every remaining convolutional block. Our FER model gives a similar performance to the state-of-the-art under a similar experimental setting, as described in \citeA{pramerdorfer2016facial}; this is higher than reported human performance \cite{goodfellow2013challenges}. 


From the FER model, we extract two classes of facial expression features to use in our image captioning models. The first class of features is the output of the final softmax layer of our FER model, $a_i = (a_{i,1}, \ldots, a_{i,7})$, representing the probability distribution of the facial expression classes for the $i$th face in the image. For the image as a whole, we construct a vector of facial expression features $s = \{s_1, \ldots, s_7\}$ used in our image captioning model as in Equation~\ref{equation:aggregate_facial_expressions}.


\begin{equation}
s_k = \begin{cases}
1\ \ \ \textnormal{for}\   k=\arg \max \sum_{1 \leq i \leq n}{a_{i,j}},\\
0\ \ \ \textnormal{otherwise}
\end{cases}
\label{equation:aggregate_facial_expressions}
\end{equation}

\noindent
where $n$ is the number of faces in the image. That is, $s$ is a one-hot encoding, which we refer to as the facial encoding vector, of the aggregate facial expressions of the image.

The second class of features consist of convolutional features extracted from the FER model, giving a more fine-grained representation of the faces in the image. For each face in an image, we extract the last convolutional layer of the model, giving $6 \times 6 \times 512$ features. We convert these into a $36 \times 512$ representation for each face. We restrict ourselves to a maximum of three faces: in our FlickrFace11K dataset, $96.5\%$ of the images have at most three faces. If one image has more than three faces, we select the three faces with the biggest bounding box sizes. We then concatenate the features of the three faces leading to $108 \times 512$ dimensions, $f=\{f_{1},...,f_{K^{\star}}\}, f_{i} \in \mathbb{R}^D$, where $K^{\star}$ is $108$ and $D$ is $512$; we refer to these as facial features. If a sample includes fewer than three faces, we fill in dimensions with zero values.

\FaceAttendF FER information. In addition to this, for comparison we trained two FER models using the high-performing ResNet and Inception architectures \cite{szegedy2015going,he2016deep}. The performance of these two additional models is similar to the VGG architecture.  Similar to the VGG-based model, we can use these ResNet and Inception-based models to extract FER features.
We use these primarily for comparison within our \FaceAttend models to assess the effect of different fine-grained facial representations. 


\subsection{Image Captioning Models}
\label{sec:imageCapModels}


Our image captioning models aim to generate an image caption, $x=\{ x_1, \dots, x_T \}$, where $x_i$ is a word and $T$ is the length of the caption, using facial expression analyses.
As a representation of the image,
all our models use the last convolutional layer of VGG-E architecture~\cite{simonyan2014very}. In addition to our proposed facial features, the VGG-E network trained on ImageNet \cite{russakovsky2015imagenet} produces a $14\times14\times512$ feature map. We convert this into a $196\times512$ representation, $c=\{c_{1},...,c_{K}\}, c_{i} \in \mathbb{R}^D$, where $K$ is $196$ and $D$ is $512$; we refer to this as the visual features. The specifics of the image captioning models are explained below.

\subsubsection{\FaceCap}
\label{sec:faceCap}
These models essentially extend the 
Show-Attend-Tell architecture of \citeA{xu2015show}.
Like these models,
we use a long short-term memory (LSTM) network as our caption generator. The LSTM incorporates the emotional content of the image in the form of the facial encoding vector defined in Equation \ref{equation:aggregate_facial_expressions}. We propose two variants, \FaceCapF and \FaceCapL, that differ in terms of how the facial encoding vector is incorporated.

\paragraph{\FaceCapF} In \FaceCapF, in each time step ($t$), the LSTM uses the previous word embedded in $M$ dimensions ($w_{t-1} \in \mathbb{R}^M$ selected from an embedding matrix learned without pre-training from random initial values), the previous hidden state ($h_{t-1}$), the attention-based features ($\hat{c}_{t}$), and the facial encoding vector ($s$) to calculate input gate ($i_t$), forget gate ($f_t$), output gate ($o_t$), input modulation gate ($g_t$), memory cell ($c_t$), and hidden state ($h_t$).
\begin{equation}
\begin{split}
& i_t = \sigma(W_{i}w_{t-1} + U_{i}h_{t-1} + C_{i}\hat{c_{t}} + S_{i}s + b_i) \\
& f_t = \sigma(W_{f}w_{t-1} + U_{f}h_{t-1} + C_{f}\hat{c_{t}} + S_{f}s + b_f) \\
& o_t = \sigma(W_{o}w_{t-1} + U_{o}h_{t-1} + C_{o}\hat{c_{t}} + S_{o}s + b_o) \\
& g_t = \tanh(W_{g}w_{t-1} + U_{g}h_{t-1} + C_{g}\hat{c_{t}} + S_{g}s + b_g) \\
& c_t = f_{t}c_{t-1}+i_{t}g_t \\
& h_t = o_t\tanh(c_t) \quad
\end{split}
\label{equarion:detail}
\end{equation} where $W, U, C, S$, and $b$ are learned weights and biases and $\sigma$ is the logistic sigmoid activation function. From now on, we show this LSTM equation using the shorthand of Equation~\ref{equation:LSTM_1}.
\begin{equation}
h_{t} = \LSTM(h_{t-1}, [\hat{c}_t, w_{t-1}, s])
\label{equation:LSTM_1}
\end{equation}

To calculate $\hat{c}_{t}$, for each time step $t$, \FaceCapF weights visual features ($c$) using a soft attention mechanism as in Equation \ref{equation:attention_first} and \ref{equation:attention_comb_first}.
\begin{equation}
\begin{split}
& e_{i,t} = W_{e}^{T} \tanh(W_{c}c_i+W_{h}h_{t-1}) \\
& e_{t}^{\prime} = \softmax(e_{t}) \quad
\end{split}
\label{equation:attention_first}
\end{equation}

\noindent
where $e_{i,t}$ are unnormalized weights for the visual features ($c_i$) and $e_{t}^{\prime}$ are the normalized weights using a softmax layer at time step $t$. Our trained weights are represented by $W_x$. Finally, our attention-based features ($\hat{c}_t$) are calculated using:
\begin{equation}
\hat{c}_t = \sum_{1 \leq i \leq K} e_{i,t}^{\prime}c_i
\label{equation:attention_comb_first}
\end{equation}

To initialize the LSTM's hidden state ($h_0$), we feed the facial features through a standard multilayer perceptron, shown in Equation \ref{equarion:LSTM_initial_1}.
\begin{equation}
h_0=\MLP_{init}( s )
\label{equarion:LSTM_initial_1}
\end{equation}

\noindent
We use the current hidden state ($h_{t}$) to calculate the negative log-likelihood of $s$ in each time step (Equation \ref{equation:face_objective}); we call this the face objective function.
\begin{equation}
L_{f}=-\sum_{1 \leq i \leq 7} s_i \log(p_e(i|h_{t})) \quad
\label{equation:face_objective}
\end{equation} 
where a multilayer perceptron generates $p_e(i|h_{t})$, which is the categorical probability distribution of the current hidden state across the facial expression classes.
We adapt this from \citeA{you2018image}, who use this objective function for injecting ternary-valued sentiment (positive, neutral, negative) into captions. This loss is estimated and averaged, over all steps, during the training phase. 

The general objective function of \FaceCapF is defined as:
\begin{equation}
L_{g1} = -\sum_{1 \leq t \leq T} \log ( p_{x}(x_t \, | \, \hat{c}_t, h_{t}) ) +
\sum_{1 \leq k \leq K}(1-\sum_{1 \leq t \leq T}c_{t})^2
\label{equation:MLE_general}
\end{equation}

\noindent
A multilayer perceptron and a softmax layer is used to calculate $p_{x}$, the probability of the next generated word:
\begin{equation}
\begin{split}
& p_{x}(x_t \, | \, \hat{c}_t, h_{t}) = {\softmax(W_c^{\prime}\hat{c}_{t} + W_{h}^{\prime}h_{t} + b^{\prime})}
\end{split}
\label{equation:MLE_general_prob}
\end{equation}

\noindent
where the learned weights and bias are given by $W^{\prime}$ and $b^{\prime}$. The last term in Equation \ref{equation:MLE_general} is to encourage \FaceCapF to equally pay attention to different sets of $c$ when a caption generation process is finished. 

\begin{figure}
	\centering
	\includegraphics[width=0.5\textwidth]{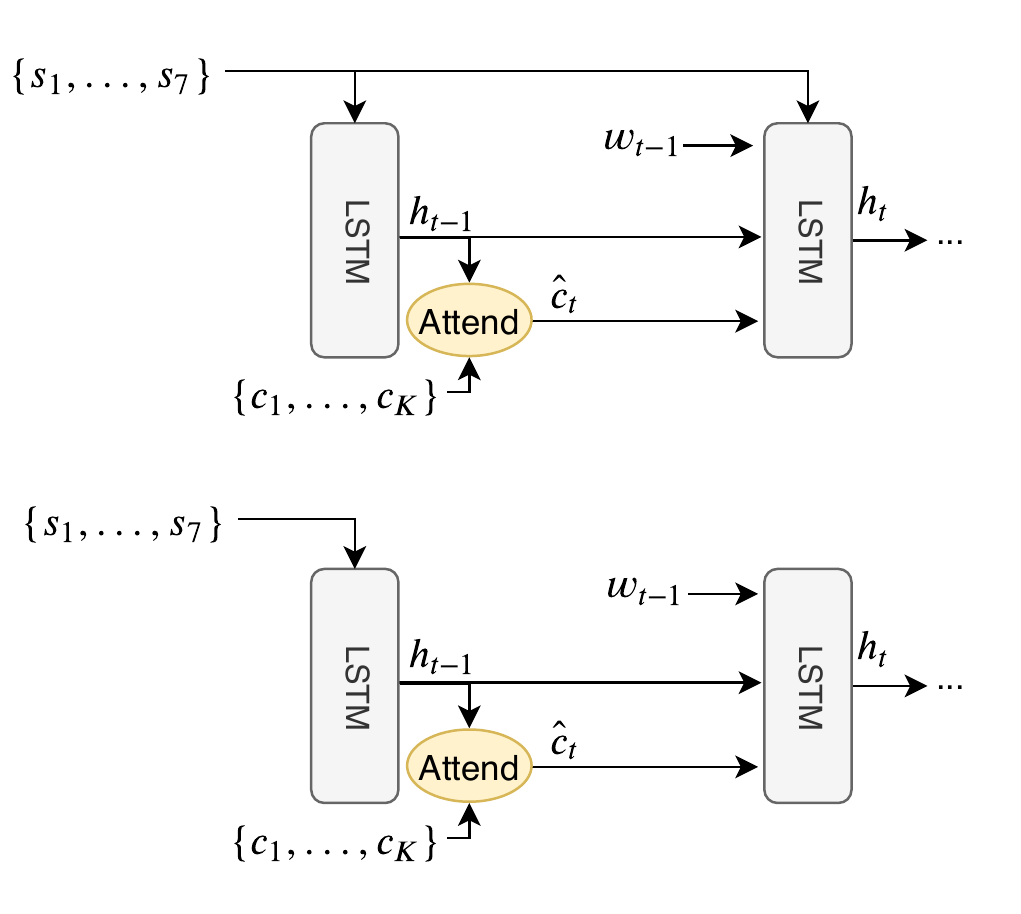}
	\caption{
		The frameworks of \FaceCapF (top), and \FaceCapL (bottom). Attend is our attention mechanism attending to the visual features, $\{c_1,\dots,c_K\}$.
	}
	\label{fig:arch_1}
\end{figure}

\paragraph{\FaceCapL} The above \FaceCapF model feeds in the facial encoding vector at the initial step (Equation \ref{equarion:LSTM_initial_1}) and at each time step (Equation \ref{equation:LSTM_1}), shown in Figure~\ref{fig:arch_1} (top). The LSTM uses the vector for generating every word because the vector is fed at each time step. Since not all words in the ground truth captions will be related to the vector --- for example in Figure~\ref{fig:example}, where the majority of words are not directly related to the facial expressions --- this mechanism could lead to an overemphasis on these features.

Our second variant of the model, \FaceCapL, is as above except that the $s$ term is removed from Equation \ref{equation:LSTM_1}: we do not apply the facial encoding vector at each time step (Figure \ref{fig:arch_1} (bottom)) and rely only on Equation \ref{equation:face_objective} to memorize this facial expression information. Using this mechanism, the LSTM can effectively take the information in generating image captions and ignore the information when it is irrelevant. To handle an analogous issue for sentiment, \citeA{you2018image} implemented a sentiment cell, working similarly to the memory cell in the LSTM, initialized by the ternary sentiment. They then fed the visual features to initialize the memory cell and hidden state of the LSTM. Similarly, \FaceCapL uses the facial features to initialize the memory cell and hidden state. Using the attention mechanism, our model applies the  visual features in generating every caption word.



\subsubsection{\FaceAttend}
\label{sec:faceAttend}
Here, we apply two LSTMs to attend to our more fine-grained facial features ({$f$}) explained in Section~\ref{sec:FER_model}, in addition to our visual features ($c$). We propose two variant architectures for combining these features, \FaceAttendF and \FaceAttendL, explained below.

\paragraph{\FaceAttendF}

\begin{figure}[t!]
  \centering
  \includegraphics[width=0.4\textwidth]{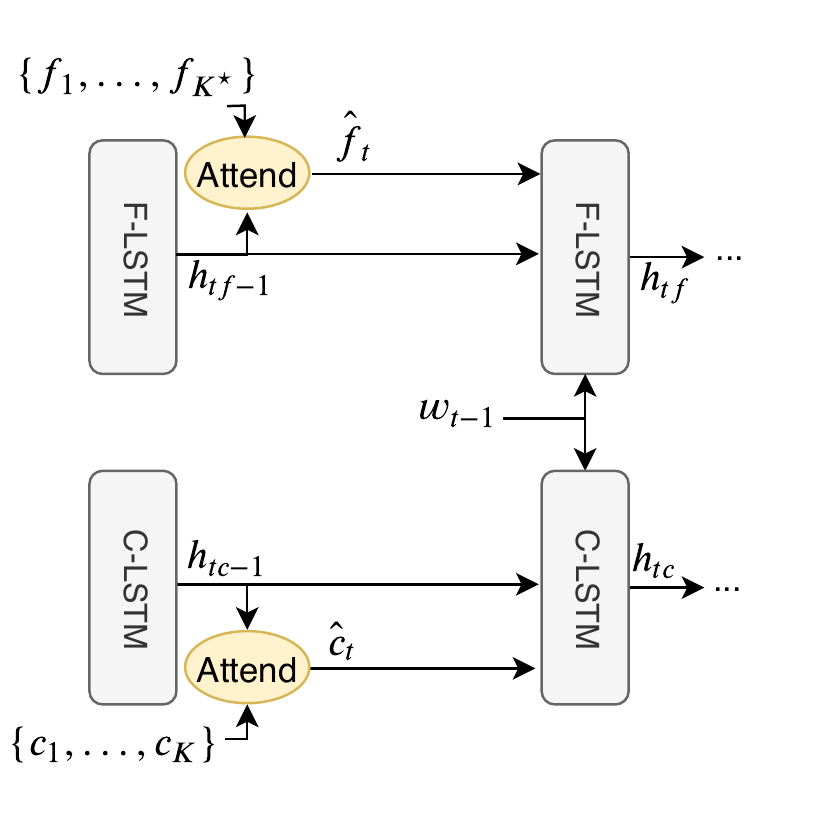}
  \caption{\FaceAttendF model enables generating image captions with both facial features $\{f_1,\dots,f_{K^{\star}}\}$ and visual content $\{c_1,\dots,c_K\}$.}
  \label{fig:arch_2}
\end{figure}

The framework of \FaceAttendF is shown in Figure~\ref{fig:arch_2}. To generate image captions, \FaceAttendF includes two LSTMs: one, called F-LSTM, to attend to facial features and another one, called C-LSTM, to attend to  visual content. Both LSTMs are defined as in Equation~\ref{equation:LSTM_2}, but with separate training parameters.
\begin{equation}
h_{t,z} = \LSTM(h_{t,z-1}, [\hat{z}_t, w_{t-1}])
\label{equation:LSTM_2}
\end{equation}

\noindent
In both LSTMs, to calculate $\hat{z}_t$ at each time step ($t$), features $z$ (the facial features ($f$) for F-LSTM and the visual features ($c$) for C-LSTM) are weighted using a soft attention mechanism, but with separately learned parameters.
\begin{equation}
\begin{split}
& e_{i,t,z} = W_{e,z}^{T} \tanh(W_{z}z_i+W_{h,z} h_{t,z-1}) \\
& e_{t,z}^{\prime} = \softmax(e_{t,z}) \quad
\end{split}
\label{equation:attention}
\end{equation}

\noindent
where $e_{i,t,z}$ and $e_{t,z}^{\prime}$ are unnormalized weights for features $z_i$, and normalized weights using a softmax layer, respectively. Our trained weights are $W_{z}$. Finally, our attention-based features ($\hat{z}_t$) are calculated using:
\begin{equation}
\hat{z}_t = \sum_{1 \leq i \leq K_z} e_{i,t,z}^{\prime}z_i
\label{equation:attention_comb}
\end{equation}

\noindent
$K_z$ is $K^{\star}$ for F-LSTM and $K$ for C-LSTM. The initial LSTM's hidden state ($h_{0,z}$) is computed using a standard multilayer perceptron:
\begin{equation}
h_{0,z}=\MLP_{init, z}( \frac{1}{K_z} \sum_{1 \leq i \leq K_z}{z_i} )
\label{equarion:LSTM_initial_2}
\end{equation}

The objective function of \FaceAttendF is defined using Equation (\ref{equation:MLE}).
\begin{multline}
L_{g2} = - \lambda [ \sum_{1 \leq t \leq T} \log(p_{x,c}(x_t \, | \, \hat{c}_t, h_{t,c}) ) + \sum_{1 \leq k \leq K}(1-\sum_{1 \leq t \leq T}c_{t,k})^2 ]  - \\ (1-\lambda) [\sum_{1 \leq t \leq T} \log(p_{x,f}(x_t \, | \, \hat{f}_t, h_{t,f}) ) + \beta_1 \sum_{1 \leq k \leq K^*}(1-\sum_{1 \leq t \leq T}f_{t,k})^2 ]
\label{equation:MLE}
\end{multline}

\vspace{5mm}

\noindent
where a multilayer perceptron and a softmax layer, for each LSTM, are used to calculate $p_{x,f}$ and $p_{x,c}$ (the probabilities of the next generated word on the basis of facial expression features and visual features, respectively):
\begin{equation}
\begin{split}
& p_{x,f}(x_t \, | \, \hat{f}_t, h_{t,f}) = {\softmax(W_f \hat{f}_{t} + W_{h,f}h_{t,f} + b_f)} \\
& p_{x,c}(x_t \, | \, \hat{c}_t, h_{t,c}) = {\softmax(W_c \hat{c}_{t} + W_{h,c}h_{t,c} + b_c)} \\
\end{split}
\label{equation:MLE_prob}
\end{equation}

\noindent
$\lambda$ and $\beta_1$ are regularization constants. The ultimate probability of the next generated word is:
\begin{equation}
\begin{split}
p_{x}(x_t \, | \, \hat{f}_t, h_{t,f}, \hat{c}_t, h_{t,c}) = \lambda p_{x,f}(x_t \, | \, \hat{f}_t, h_{t,f}) +
 (1-\lambda) p_{x,c}(x_t \, | \, \hat{c}_t, h_{t,c})
\end{split}
\label{equation:word_predict}
\end{equation}

\paragraph{\FaceAttendL}

\begin{figure}[t!]
  \centering
  \includegraphics[width=0.55\textwidth]{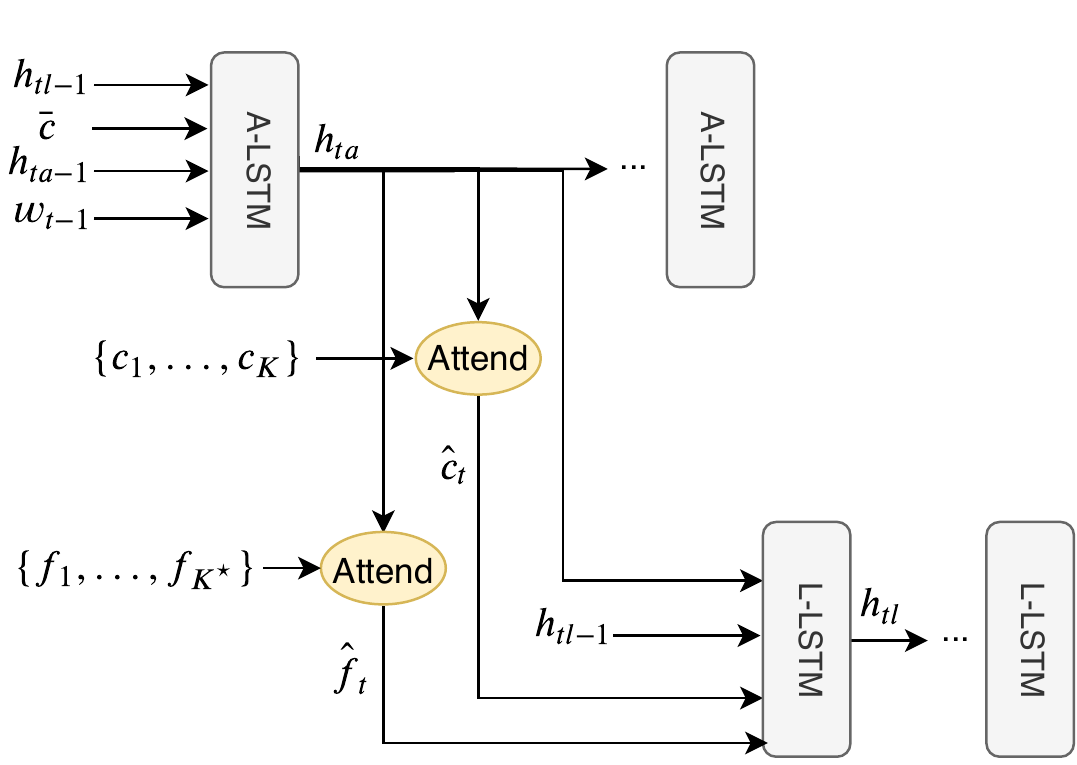}
  \caption{\FaceAttendL model enables generating image captions with two LSTMs for learning attention weights and generating captions, separately.}
  \label{fig:arch_3}
\end{figure}

The above \FaceAttendF model uses two LSTMs: one for attending to visual features and another one for attending to facial features. In the model, both LSTMs also play the role of language models (Equation \ref{equation:word_predict}) and directly impact on the prediction of the next generated word. However, the recent state-of-the-art image captioning model of \citeA{anderson2018bottom} achieved  better performance by using two LSTMs with differentiated roles: one for attending only to visual features and a second one purely as a language model. Inspired by this, we define our \FaceAttendL variant to use one LSTM, which we call A-LSTM, to attend to image-based features, both facial and visual; and a second one, which we call L-LSTM, to generate language (Figure \ref{fig:arch_3}). Here, we calculate the hidden state of A-LSTM using:
\begin{equation}
h_{t,a} = \LSTM(h_{t,a-1}, [\bar{c}, h_{t,l-1}, w_{t-1}])
\label{equation:ALSTM}
\end{equation}

\noindent
where $\bar{c} = \frac{1}{K} \sum_{1 \leq i \leq K}{c_i}$ is the mean-pooled visual features and $h_{t,l-1}$ is the previous hidden state of L-LSTM. We also calculate the hidden state of L-LSTM using:
\begin{equation}
h_{t,l} = \LSTM(h_{t,l-1}, [\hat{f}_t,\hat{c}_t, h_{t,a}])
\label{equation:LLSTM}
\end{equation}

\noindent
where $\hat{f}_t$ and $\hat{c}_t$ are the attended facial features and visual features, respectively. They are defined analogously to Equation \ref{equation:attention} and \ref{equation:attention_comb}, but $h_{t,z-1} = h_{t,a}$ with different sets of trainable parameters. $h_a$ and $h_l$ are similarly initialized as follows using two standard multilayer perceptrons:
\begin{equation}
\begin{split}
& h_{0,l} =\MLP_{init,l}( \frac{1}{K} \sum_{1 \leq i \leq K}{c_i} ) \\
& h_{0,a}=\MLP_{init,a}( \frac{1}{K} \sum_{1 \leq i \leq K}{c_i} )
\end{split}
\label{equarion:LSTM_initial_3}
\end{equation}

The objective function of \FaceAttendL is:
\begin{equation}
L_{g3} = - [\sum_{1 \leq t \leq T} \log( p_{x}(x_t \, | \, \hat{c}_t, \hat{f}_t, h_{t,l}) )
+
 \sum_{1 \leq k \leq K}(1-\sum_{1 \leq t \leq T}c_{t,k})^2 +  \beta_2 \sum_{1 \leq k \leq K^{\star}}(1-\sum_{1 \leq t \leq T}f_{t,k})^2]
\label{equation:MLE_topdown}
\end{equation}

\noindent
where 
$\beta_2$ is a regularization constant and $p_{x}$ is the probability of the next generated word calculated as follows:
\begin{equation}
\begin{split}
& p_{x}(x_t \, | \, \hat{c}_t, \hat{f}_t, h_{t,l}) = {\softmax( W_{c,l}\hat{c}_{t} + W_{f,l}\hat{f}_{t} + W_{h,l}h_{t,l} + b_l)} \\
\end{split}
\label{equation:MLE_prob_topdown}
\end{equation}

\noindent
where $W_{x,l}$ and $b_l$ are trainable weights and bias, respectively.


\section{Experiments}
\label{sec:exper}

\subsection{Evaluation Metrics} 
Following previous work, we evaluate our image captioning model using standard evaluation metrics including BLEU~\cite{papineni2002bleu}, ROUGE~\cite{lin2004rouge}, METEOR~\cite{denkowski2014meteor}, CIDEr~\cite{vedantam2015cider}, and SPICE~\cite{anderson2016spice}. Larger values are better results for all metrics. BLEU calculates a weighted average for n-grams with different sizes as a precision metric. ROUGE is a recall-oriented metric that calculates F-measures using the matched n-grams between the generated captions and their corresponding reference summaries. METEOR uses a weighted F-measure matching synonyms and stems in addition to standard n-gram matching. CIDEr uses a n-gram matching, calculated using the cosine similarity, between the generated captions and the consensus of the reference captions. Finally, SPICE calculates F-score for semantic tuples derived from scene graphs.


\subsection{Systems for Comparison} 

The core architectures for our \FaceCap and \FaceAttend models come from Show-Attend-Tell \cite{xu2015show} and Up-Down-Captioner \cite{anderson2018bottom}, respectively.
We therefore use these models, trained on the FlickrFace11K dataset, as baselines, in order to provide an ablative assessment of the effect of adding facial expression information. We call these baseline models \XU and \ANDERSON. (Moreover, \citeA{anderson2018bottom} has the state-of-the-art results for image captioning.)

We further look at two additional models to investigate the impact of the face loss function in using the facial encoding in different schemes.
We train the \FaceCapF model, which uses the facial encoding in every time step, without calculating the face loss function (Equation (\ref{equation:face_objective})); we refer to this (following the terminology of \citeA{hu2017toward} and \citeA{you2018image})
as the \FaceStep model. The \FaceCapL model, which applies the facial encoding in the initial time step, is also modified in the same way; we refer to this as the \FaceInit model. 

\subsection{Implementation Details}
The size of the word embedding layer, initialized via a uniform distribution, is set to $300$ except for \ANDERSON and \FaceAttendL which is set to 512. We fixed $512$ dimensions for the memory cell and the hidden state in this work. We use the mini-batch size of $100$ and the initial learning rate of $0.001$ to train each image captioning model except \ANDERSON and \FaceAttendL where we set the mini-batch size to 64 and the initial learning rate to $0.005$. We used different parameters for \ANDERSON and \FaceAttendL in comparison with other models because using similar parameters led to worse results for all models. The Adam optimization algorithm~\cite{kingma2014adam} is used for optimizing all models. During the training phase, if the model does not have an improvement in METEOR score on the validation set in two successive epochs, we divide the learning rate by two (the minimum learning rate is set to $0.0001$) and the previous trained model with the best METEOR is reloaded. This method of learning rate decay is inspired by \citeA{wilson2017marginal}, who advocated tuning the learning rate decay for Adam. In addition to learning rate decay, METEOR is applied to select the best model on the validation set because of a reasonable correlation between METEOR and human judgments~\cite{anderson2016spice}. Although SPICE can have higher correlations with human judgements, METEOR is quicker to calculate than SPICE, which requires dependency parsing, and so more suitable for a training criterion. The epoch limit is set to 30. We use the same vocabulary size and visual features for all models. $\lambda$ and $\beta_1$ in Equation \ref{equation:MLE} are empirically set to $0.8$ and $0.2$, respectively. $\beta_2$ in Equation \ref{equation:MLE_topdown} is also set to $0.4$.
Multilayer perceptrons in Equation~\ref{equarion:LSTM_initial_1}, {\ref{equarion:LSTM_initial_2}} and {\ref{equarion:LSTM_initial_3}} use $\tanh$ as an activation function.

\subsection{Experimental Results}
\paragraph{Quantitative Analysis: Performance Metrics}
The FlickrFace11K splits are used for training and evaluating all image captioning models in this paper. Table~\ref{tab:results} summarizes the results on the FlickrFace11K test set. \FaceAttendF and \FaceAttendL outperform other image captioning models using all the evaluation metrics. For example, \FaceAttendF achieves 17.6 for BLEU-4 which is 1.9 and 0.4 points better that \XU (the first baseline model) and \FaceCapL (the best of the \FaceCap models), respectively. \FaceAttendL also achieves a BLEU-4 score of 17.7 which is 0.4 better than \ANDERSON, the baseline model it builds on, and 0.5 better than \FaceCapL. 
\FaceAttendF and \FaceAttendL show very close results, with \FaceAttendF demonstrating a couple of larger gaps in performance, in the BLEU-1 and ROUGE-L metrics.  Among the \FaceCap models, \FaceCapL is clearly the best. 

\green{Table~\ref{tab:results_diff_arch} compares \FaceAttendF-VGG with FER features derived from the VGG architecture (\FaceAttendF in Table~\ref{tab:results} which is our core version in the paper) against \FaceAttendF-RES using the ResNet architecture and \FaceAttendF-INC using the Inception architecture (see Section~\ref{sec:FER_model}). This comparison is to investigate the variability of FER features derived from different architectures on the image captioning task; we choose  \FaceAttendF for this as the highest-performing model from Table~\ref{tab:results}.  All three \FaceAttendF in the table perform similarly, and outperform the \XU model, using all the image captioning metrics. This confirms the broadly similar effectiveness of the FER features from different architectures.}


\begin{table}
\caption{The results of different image captioning models (\%) on FlickrFace11K test split. B-N is the BLEU-N metric. The best performances are bold.}\label{tab:results}
\centering
\resizebox{\textwidth}{!}{
\begin{tabular}{|l|c|c|c|c|c|c|c|c|}
\hline
\textbf{Model} & \textbf{B-1} & \textbf{B-2} & \textbf{B-3} & \textbf{B-4} & \textbf{METEOR} & \textbf{ROUGE-L} & \textbf{CIDEr} & \textbf{SPICE} \\
\hline
\hline
\XU & 56.0  & 35.4 & 23.1 & 15.7 & 17.0 & 43.7 & 21.9 & 9.3 \\
\hline
\ANDERSON & 57.9 & 37.3 & 25.0 & 17.3 & 17.5 & 45.1 & 24.4 & 10.1 \\
\hline
\hline
\FaceStep & 58.4  & 37.6 & 24.8 & 17.0 & 17.5 & 45.0 & 22.8 & 9.9 \\
\hline
\FaceInit & 56.6  & 36.5 & 24.3 & 16.9 & 17.2 & 44.8 & 23.1 & 9.8 \\
\hline
\FaceCapF & 57.1 & 36.5 & 24.1 & 16.5 & 17.2 & 44.8 & 23.0 & 9.7 \\
\hline
\FaceCapL & 58.9 & 37.9 & 25.1 & 17.2 & 17.4 & 45.5 & 24.7 & 10.0  \\
\hline
\hline
\FaceAttendF & \textbf{59.4} & \textbf{38.2} & 25.4 & 17.6 & \textbf{17.6} & \textbf{45.8} & \textbf{24.9} & 10.1  \\
\hline
\FaceAttendL & 58.6 & 38.1 & \textbf{25.6} & \textbf{17.7} & \textbf{17.6} & 45.5 & 24.8 & \textbf{10.2}  \\
\hline
\end{tabular}}
\end{table}

\begin{table}
\caption{\green{\FaceAttendF with different sets of FER features, extracted by our FER models using high-performing CNN architectures including VGG, ResNet (RES) and Inception (INC).}}\label{tab:results_diff_arch}
\centering
\resizebox{\textwidth}{!}{
\begin{tabular}{|l|c|c|c|c|c|c|c|c|}
\hline
\textbf{Model} & \textbf{B-1} & \textbf{B-2} & \textbf{B-3} & \textbf{B-4} & \textbf{METEOR} & \textbf{ROUGE-L} & \textbf{CIDEr} & \textbf{SPICE} \\
\hline
\hline
\FaceAttendF - VGG & 59.4 & 38.2 & 25.4 & 17.6 & 17.6 & 45.8 & 24.9 & 10.1  \\
\FaceAttendF - RES & 58.8 & 38.1  & 25.4 & 17.6 & 17.3  & 45.3 & 23.4  & 9.8  \\
\FaceAttendF - INC & 58.7 & 38.0 & 25.4 & 17.7 & 17.3 & 45.3 & 23.5 & 9.7  \\
\hline
\end{tabular}}
\end{table}

\paragraph{Quantitative Analysis: Entropy, Top$_4$ and Ranking of Generated Verbs}
To analyze what it is about the captions themselves that differs under the various models, with respect to our aim of injecting information about emotional states of the faces in images, we first extracted all generated adjectives, which are tagged using the Stanford part-of-speech tagger software~\cite{toutanova2003feature}. Perhaps surprisingly, emotions do not manifest themselves in the adjectives in our models: the adjectives used by all systems are essentially the same. 

To investigate this further, we took the NRC emotion lexicon\footnote{\url{https://saifmohammad.com/WebPages/NRC-Emotion-Lexicon.htm}} \cite{mohammad-turney:2013} and examined the occurrence of words in the captions that also appeared in the lexicon.  This widely-used lexicon is characterised as ``a list of English words and their associations with eight basic emotions (anger, fear, anticipation, trust, surprise, sadness, joy, and disgust)'' whose labels have been manually annotated through crowd-sourcing.  The labels are based on word associations --- annotators were asked ``which emotions are associated with a target term'' --- rather than whether the word \emph{embodies} an emotion; the lexicon thus contains a much larger set of words than is useful for our purposes.  (For example, the most frequent word overall in the reference captions that appears in the lexicon is \textit{young}, which presumably has some positive emotional associations.)
In addition, the set of emotions used in lexicon labels does not exactly correspond to our set.  We therefore do not propose to use this lexicon purely automatically, but instead to help in understanding the use of emotion-related words.

Among the reference captions, as noted above the most frequent word from the emotion lexicon was \textit{young}, followed by \textit{white}, \textit{blue} and \textit{black}; all of these presumably have some emotional association, but do not generally embody an emotion.  The first word embodying the expression of an emotion is the verb \textit{smiling}, at rank 8, with other similar verbs following closely (e.g. \textit{laughing}, \textit{enjoying}).  The highest ranked emotion-embodying adjective is \textit{happy} at rank 26, with a frequency of around 15\% of that of \textit{smiling}; other adjectives were much further behind.  It is clear that verbs form a more significant expression of emotion in this particular dataset than do adjectives.

To come up with an overall quantification of the different linguistic properties of the generated captions under the models, we therefore focused our investigation on the differences in distributions of the generated verbs. 
To do this, we calculated three measures.
The first is entropy (in the information-theoretic sense), which can indicate which distributions are closer to deterministic and which are more spread out (with a higher score indicating more spread out): in our context, it will indicate the amount of variety in selecting verbs.
We calculated entropy using the standard Equation (\ref{equation:entropy}).

\begin{equation}
\entropy = -\sum_{1 \leq i \leq V}{p(v_i)\times\log_2(p(v_i))} \quad
\label{equation:entropy}
\end{equation}

\noindent
where $V$ indicates the number of the unique generated verbs and $p(v_i)$ is the probability of each generated verb ($v_i$), estimated as the Maximum Likelihood Estimate from the sample. 

As a second measure, we looked at the four most frequent verbs (Top$_4$), which are the same for all models (\textit{is}, \textit{sitting}, \textit{are}, \textit{standing}) --- these are verbs with relatively little semantic content, and for the most part act as syntactic props for the content words of the sentence.  The amount of probability mass left beyond those four verbs is another indicator of variety in verb expression.

\begin{table}[t]
	\caption{The Entropies of all generated verbs and the probability mass of the Top$_4$ generated verbs (\textit{is}, \textit{are}, \textit{sitting}, and \textit{standing}). Reference means the ground-truth captions.}
	\label{tab:verb-distrib}
	\centering
		\begin{tabular}{|l|c|c|}
			\hline
			\textbf{Model} & \textbf{Entropy} & \textbf{Top$_4$}\\
			\hline
			\hline
			Reference & 6.9963 & 32.63\% \\
			\hline
			\hline
			\XU & 2.7864 & 77.05\% \\
			\hline
			\ANDERSON & 2.7092 & 79.24\% \\
			\hline
			\hline
		    \FaceStep & 2.9059 & 74.80\% \\
		    \hline
			\FaceInit & 2.6792 & 78.78\%\\
			\hline
			\FaceCapF & 2.7592 & 77.68\% \\
			\hline
			\FaceCapL & 2.9306 & 73.65\% \\
			\hline
			\hline
			\FaceAttendF & 3.0154 & 71.14\% \\
			\hline
			\FaceAttendL & 2.8074 & 77.69\% \\
			\hline
		\end{tabular}
\end{table}

\begin{table}[t]
	\caption{Comparison of different image captioning models in ranking example generated verbs. Higher ranks mean better results.}
	\label{tab:verb-examples}
	\begin{center}
		\begin{tabular}{|l|c|c|c|c|c|c|c|c|}
			\hline
			\textbf{Model} & \textbf{Smiling} & \textbf{Looking} & \textbf{Singing} & \textbf{Reading} & \textbf{Eating} & \textbf{Laughing} \\
			\hline
			\hline
			Reference & 11 & 10 & 27 & 35 & 24 & 40 \\
			\hline
			\hline
			\XU & 19 & n/a & 15 & n/a & 24 & n/a \\
			\hline
			\ANDERSON & 14 & 13 & 9 & n/a & 15 & n/a \\
			\hline
			\hline
			\FaceStep & 11 & 18 & 10 & n/a & 15 & n/a\\
			\hline
			\FaceInit & 10 & 21 & 12 & n/a & 14 & n/a\\
			\hline
			\FaceCapF & 12 & 20 & 9 & n/a & 14 & n/a \\
			\hline
			\FaceCapL & 9 & 18 & 15 & 22 & 13 & 27 \\
			\hline
			\hline
			\FaceAttendF & 14 & 16 & 9 & 19 & 19 & 25 \\
			\hline
			\FaceAttendL & 15 & 13 & 8 & 15 & 17 & 23 \\
			\hline
		\end{tabular}
	\end{center}
\end{table}

Table~\ref{tab:verb-distrib} shows that \FaceAttendF can generate the most diverse distribution of the verbs compared to other models because it has the highest Entropy. It also shows that \FaceAttendF has the lowest (best) proportion of the probability mass taken up by Top$_4$, leaving more for other verbs. In contrast to the results of the standard image captioning metrics shown in Table \ref{tab:results}, \FaceAttendF and \FaceAttendL show very different behaviour: \FaceAttendF is clearly superior.  Among the \FaceCap models, as for the overall metrics, \FaceCapL is the best, and is in fact better than \FaceAttendL.
(As a comparison, we also show Entropy and Top$_4$ for all reference captions (5 human-generated captions per image): human-generated captions are still much more diverse than the best models.)



The two measures above are concerned only with variety of verb choice and not with verbs linked specifically to emotions or facial expressions.
For a third measure, therefore, we look at selected individual verbs linked to actions that relate to facial emotion expression, either direct or indirect.  Our measure is the rank of the selected verb among all those chosen by a model; higher (i.e. lower-numbered) ranked verbs mean that the model more strongly prefers this verb.  Our selected verbs are among those that ranked highly in the reference captions and also appeared in the emotion lexicon.

Table~\ref{tab:verb-examples} shows a sample of those verbs such as  \textit{singing}, \textit{reading} and \textit{laughing}.
The baseline \XU model ranks all of those relatively low, where our other baseline \ANDERSON and our models incorporating facial expressions do better.  Only \FaceCapL (the best of our \FaceCap models by overall metrics) and our \FaceAttend models manage to use verbs like \textit{laughing} and \textit{reading}.

\begin{figure}
  \centering
  \includegraphics[width=0.75\textwidth]{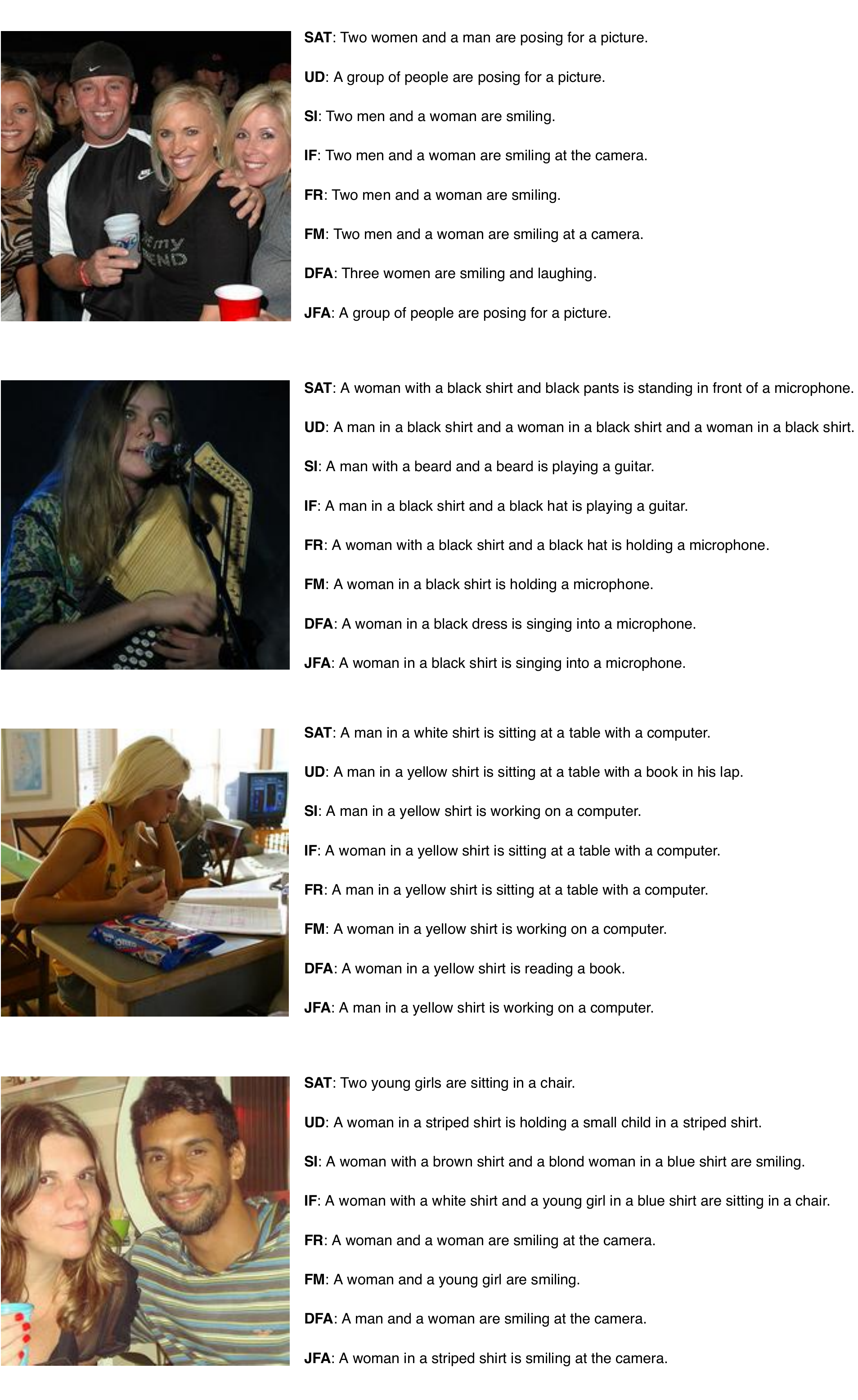}
  \caption{Example generated captions using SAT (\XU), UD (\ANDERSON) SI (\FaceStep), IF (\FaceInit), FR (\FaceCapF), FM (\FaceCapL), DFA (\FaceAttendF) and JFA (\FaceAttendL) models.}
  \label{fig:cap_out}
\end{figure}

\paragraph{Qualitative Analysis}
In Figure~\ref{fig:cap_out}, we compare some generated captions by different image captioning models using four representative images. The first one shows that \FaceAttendF correctly uses \textit{smiling} and \textit{laughing} to capture the emotional content of the image. \FaceStep, \FaceInit, \FaceCapF and \FaceCapL are also successful in generating \textit{smiling} for the image. For the second sample, \FaceAttendF and \FaceAttendL use the relevant verb  \textit{singing} to describe the image, while other models cannot generate the verb. Similarly, \FaceAttendF generates the verb \textit{reading} for the third image. Moreover, most models can correctly generate \textit{smiling} for the forth image except \XU and \ANDERSON which do not use the facial information. \FaceInit also cannot generate \textit{smiling} because it uses the facial information only at initial step which provides a weak emotional signal for the model. Here, \FaceAttendF can generate the most accurate caption (``A man and a woman are smiling at the camera'') for the image, while other models generate some errors. For example, \FaceCapL generates ``A woman and a young girl are smiling'', which does not describe the man in the image.

\subsection{\green{Failure Analyses}}
\green{We also carried out an analysis on examples where our image captioning models fail to generate better captions than the baseline models. We first look quantitatively at these examples via image captioning metrics, focussing on SPICE, and then show a few of these examples.}

\paragraph{\green{Quantitative Analysis: SPICE}}
For our failure analysis, we use the SPICE metric to compare generated captions by different models: SPICE is specifically designed for fine-grained analyses, as described in \citeA{anderson2016spice}, as it can break down scoring 
into semantic proposition subcategories including object, relation, and attribute; it can also break down attributes further into color, count and size, for example.

To identify examples where our image captioning models perform worse, we first calculate SPICE scores on individual examples.  As image captioning metrics are designed to be applied to a set of captions rather than individual ones, this only gives a rough idea of the quality of an individual caption; we therefore set a threshold on the difference between our models and the baseline (0.05) so as not to include ones where scores are very close and therefore may not be a reliable indicator that the caption is actually worse.

Our analysis uses \XU as the baseline model without FER features, and two of our models: \FaceCapL (our best version using the FER one-hot encoding), and \FaceAttendF (our best version using the FER convolutional features). 

We first show the SPICE F-scores for subcategories over \textit{all} captions, in Table~\ref{tab:results_subcategory_all}.  We observe in general that although the overall SPICE scores for our models are better (as in Table~\ref{tab:results}), they are lower for the size attribute, showing that adding facial expression features can reduce the focus on this attribute in describing visual content. This is particularly the case for \FaceCapL which uses the one-hot encoding version of the features. 
\FaceCapL is similar to \XU but worse than \FaceAttendF in terms of the count attribute, perhaps because the one-hot encoding here presents the aggregate facial expressions of the input image (Section~\ref{sec:FER_model}) and ignores the number of individuals in the image.

In terms of the selected subset of captions where our models perform worse, Table~\ref{tab:results_subcategory_lower} shows the average SPICE F-scores for these subcategories.  The key difference here is that \XU performs a lot better on this subset in terms of overall SPICE score than it does on all captions (15.6 vs 9.3) while our two models perform just slightly worse on these than on all captions (\FaceCapL: 8.7 vs 10.0; \FaceAttendF: 9.3 vs 10.1).  This relationship also holds for the object subcategory, and for \XU and \FaceCapL for the relation category.  This may be because the FER features encourage the models to generate relevant verbs (e.g., \textit{smiling}, \textit{looking}) and nouns (e.g. \textit{camera}, \textit{microphone}) as shown in Table~\ref{tab:verb-examples} and Figure~\ref{fig:cap_out}, which are sometimes less relevant.  Overall, the trade-off that our models appear to make is that performance degrades slightly on images that they are less well-suited to, while boosting performance overall.

\begin{table}
\caption{\green{SPICE and F-scores of the semantic subcategories for all captions generated using different models.}}\label{tab:results_subcategory_all}
\centering
\resizebox{\textwidth}{!}{
\begin{tabular}{|l|c|c|c|c|c|c|c|c|}
\hline
\textbf{Model} & \textbf{SPICE} & \textbf{Object} & \textbf{Relation} & \textbf{Attribute} & \textbf{Color} & \textbf{Count} & \textbf{Size} \\
\hline
\hline
\XU & 9.3 & 19.4 & 3.0 & 3.7 & 8.0 & 2.3 & 2.9   \\
\FaceCapL & 10.0 & 20.5  & 3.1 & 4.5 & 10.1  & 2.4 & 1.3    \\
\FaceAttendF & 10.1 & 20.1 & 3.2 & 4.5 & 10.1 & 4.1 & 2.3  \\
\hline
\end{tabular}}
\end{table}

\begin{table}
\caption{\green{SPICE and F-scores of the semantic subcategories where our models generate captions with lower scores compared to the baseline model.}}\label{tab:results_subcategory_lower}
\centering
\resizebox{\textwidth}{!}{
\begin{tabular}{|l|c|c|c|c|c|c|c|c|}
\hline
\textbf{Model} & \textbf{SPICE} & \textbf{Object} & \textbf{Relation} & \textbf{Attribute} & \textbf{Color} & \textbf{Count} & \textbf{Size} \\
\hline
\hline
\XU & 15.6 & 30.7 & 6.4 & 6.9 & 13.9 & 6.6 & 6.5   \\
\FaceCapL & 8.7 & 18.5  & 2.0 & 4.0 & 8.7  & 3.5 & 0.8    \\
\FaceAttendF & 9.3 & 18.8 & 3.3 & 4.4 & 9.0 & 6.4 & 2.4  \\
\hline
\end{tabular}}
\end{table}

\begin{figure}[t]
  \centering
  \includegraphics[width=0.30\textwidth]{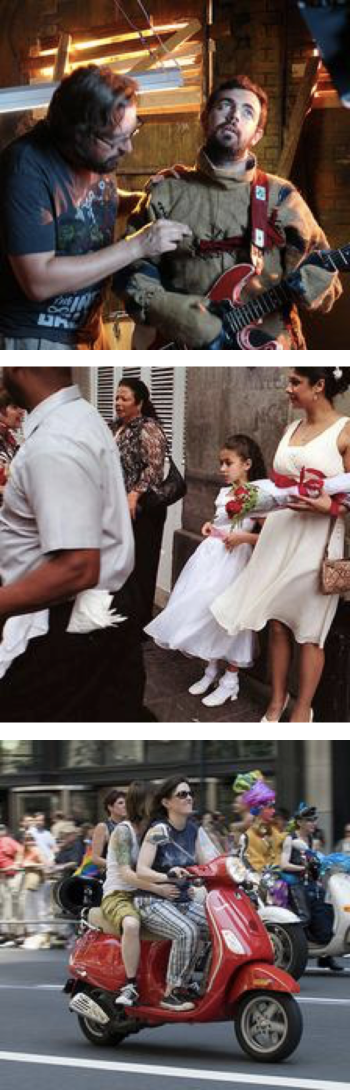}
  \caption{Example images where \FaceCap and \FaceAttendF fail to generate better results than \XU.}
  \label{fig:fail_one}
\end{figure}

\begin{figure}[t]
  \centering
  \includegraphics[width=0.3\textwidth]{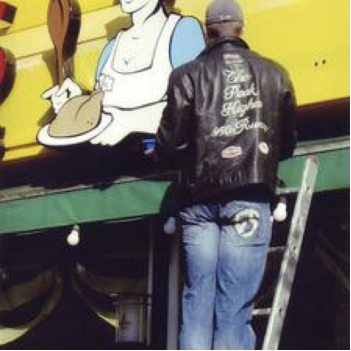}
  \caption{An example image where the \XU model expresses the size attribute in its generated caption.}
  \label{fig:fail_two}
\end{figure}

\paragraph{\green{Examples}}
Figure~\ref{fig:fail_one} shows some examples where our two models produce substantially worse captions than the baseline according to the SPICE metric.  In the topmost one 
the baseline \XU generates ``a man with a beard and a woman in a black shirt are playing a guitar'' while \FaceCapL generates ``a woman is playing a guitar and singing into a microphone'' and \FaceAttendF generates ``two men are playing a guitar and singing''.  Notwithstanding some gender confusions, the baseline is scoring higher because of the mention of the clothing, which appears in two of the human reference captions, while our models have incorrectly guessed the people are singing (perhaps not unreasonably, given the guitar); \FaceCapL also postulates the existence of a microphone.

In the middle image, \XU generates ``two women in a white dress and a man in a white shirt are standing in a crowd'' while \FaceCapL generates ``a group of people are dancing together'' and \FaceAttendF generates ``a man in a white shirt and a woman in a white shirt are standing in front of a microphone''.  This is another instance where a face-focussed model posits the existence of a microphone, as an object that is commonly near a face.

The final image appears to be one of a number of instances where SPICE is likely to be an inaccurate reflection of human judgement of the relative quality of the models.  \XU has ``a woman in a red shirt is sitting on a bench with a large large crowd on the side of'', while \FaceCapL has ``a man in a red shirt is riding a bike'' and \FaceAttendF has ``a group of people are riding bikes on a street''.  None of the human reference captions use the word ``bike'' even though that is a prominent aspect of the image (there is ``moped'' and (sic) ``mo pad''), while the less salient ``crowd'' is mentioned in one reference caption, boosting the score of \XU.

As we were also curious about the unexpected advantage that the baseline \XU has in terms of the size attribute, and noting the repeated ``large large'' generated by \XU in this last example, while conducting our failure analysis we also looked at other examples where this model expressed size.  We found that a large number of them looked like the one in Figure~\ref{fig:fail_two}, where \XU generated ``a man in a blue shirt is standing in a room with a large large large large large large large''.  This was in contrast to \FaceCapL's ``a man in a blue jacket is standing in front of a yellow wall'' and \FaceAttendF's ``a man in a blue shirt is standing in front of a green car'', which are less problematic even though the \XU model scored better than \FaceAttendF.  This repeated word problem is known from neural machine translation \cite{mi-etal:2016:EMNLP} and is common to neural models in general.  Exploring this issue is beyond the scope of this paper, but we do note based just on our observations that our models seemed less prone to this problem of repeated size attributes, even though the SPICE size attribute scores suggest the baseline \XU is evaluated at being better at describing sizes.


\section{Conclusion}
In this work, we have presented several image captioning models incorporating emotion-related information from facial features. All of our models produce better captions on images including faces than strong baseline systems, as measured by standard metrics on the FlickrFace11K dataset.  In investigating these models, we made the following findings:

\begin{itemize}
    \item Our models that use a distributed representation of facial emotion (\FaceAttend) outperformed those that use a one-hot encoding (\FaceCap).
    
    \item For \FaceCap models, injecting facial expression information only once at the start outperformed injecting at all steps in caption generation, suggesting that the models shouldn't encourage too strongly the incorporation of this facial expression information.  For \FaceAttend models, our two different methods for separating information --- separate LSTMs for visual and facial features (\FaceAttendF) versus separate LSTMs for visual and language functions (\FaceAttendL) --- performed fairly similarly in terms of overall metrics.
    
    \item
    A linguistic analysis of the generated captions showed that much of the improvement in our models was manifested through verbs.  In particular, under measures of diversity of caption generation, \FaceAttendF was substantially better than all other models.
    
    \item
    An ablative study on the distributed facial emotion representations in \FaceAttend showed similar performance regardless of which of three high-performing facial emotion recognition systems was used.
    
    \item
    A failure analysis showed only minor degradation of performance in those cases where the baseline outperformed out new models.
    
\end{itemize}


In terms of improvements to our models, the failure analysis suggests the addition of some mechanism that prevents the models from too strongly encouraging the caption generator to incorporate objects that are associated with faces; the findings regarding the location for incorporating facial expressions in the architecture (in \FaceCapL versus \FaceCapF) could be explored in our other models too.

In terms of broader application of the ideas of this work, there is other recent work that explore other aspects of emotional content in images; we note specifically the dataset of \citeA{you-etal:2016:AAAI}.
In future work, we are interested in exploring this broader emotional content of images, which is reflected in the NRC Emotion Lexicon we used in our linguistic analysis of captions.


\vskip 0.2in
\bibliography{sample}
\bibliographystyle{theapa}

\end{document}